%% file: acl_latex.tex
\pdfoutput=1
\documentclass[11pt]{article}

\usepackage[preprint]{acl}

\usepackage{times}
\usepackage{latexsym}

\usepackage[T1]{fontenc}

\usepackage[utf8]{inputenc}

\usepackage{microtype}

\usepackage{inconsolata}

\usepackage{graphicx}
\usepackage{booktabs}
\usepackage{amsmath}
\usepackage{makecell}
\usepackage{colortbl}
\usepackage{array}
\usepackage{longtable}
\usepackage{amssymb}
\usepackage{subcaption}
\usepackage{multirow} 
\usepackage{listings}
\usepackage{xcolor}
\colorlet{numb}{magenta!60!black}
\usepackage{enumitem}
\usepackage{url}

\title{Benchmarking Patent Drafting from Inventor-Style Disclosures}

\author{
Lekang Jiang$^{1}$, 
Wenjun Sun$^{1}$$^{2}$$^{3}$,
Stephan Goetz$^{1}$ \\
$^{1}$University of Cambridge $^{2}$National Science Library, Chinese Academy of Sciences \\ $^{3}$Department of Information Resources Management, School of Economics and Management,\\ University of Chinese Academy of Sciences \\
\texttt{\{lj408, ws462, smg84\}@cam.ac.uk}
}

\begin{document}
\maketitle

\begin{abstract}
While recent large language models (LLMs) have achieved promising results on individual patent drafting tasks, they fundamentally fail to investigate the core challenge of real-world patent drafting: generating a complete and legally coherent patent application directly from early-stage invention materials. Prior work predominantly assumes later-stage, highly structured, or already legalistic inputs. However, real patenting workflows begin with informal, de-legalized disclosures authored by inventors. To bridge the gap, we introduce Dis2Pat, a disclosure-to-patent dataset that reflects realistic patenting workflows by requiring the generation of complete patent applications directly from inventor-style, de-legalized disclosures. Given the inherent difficulty of long-form, legally constrained patent drafting and the strong privacy requirements, we further propose a strong baseline named Patent-MAF. It is a multi-agent framework for locally deployable patent drafting. Benchmark results reveal that current LLMs exhibit limitations in patent drafting, while Patent-MAF provides a strong baseline that consistently outperforms evaluated open-source models and remains competitive with large closed-source models. \footnote{\url{https://github.com/scylj1/Patent-MAF}}
\end{abstract}

\section{Introduction}

\input{tabs/tab_relatedwork}

Patent drafting is a critical yet costly process that transforms an inventor’s technical ideas into legally enforceable documents consisting of claims and a detailed specification \citep{epo2020}. Claims define the legal scope of a patent, while the specification provides comprehensive descriptions of the invention and its specific embodiments, typically with reference to the accompanying drawings \citep{jiang2025natural}. Although recent advances in large language models (LLMs) have achieved promising results on individual patent drafting tasks---such as claim generation \citep{jiang-etal-2025-large,jiang-etal-2025-enriching,jiang-etal-2025-patent} and specification writing \citep{wang-etal-2024-patentformer,yang2025patentvision,shea-yu-2025-autospec}---the generation of a complete patent application from early-stage invention materials remains an open problem.

A key challenge arises from the mismatch between existing benchmarks and real-world patent drafting workflows. In practice, inventors typically submit informal invention disclosures describing the problem, core idea, implementation details, and illustrative figures; patent attorneys then translate these materials into legally structured claims and specifications \citep{CisloAndThomas2023}. However, most prior datasets and models operate on later-stage or highly structured inputs, such as full specifications for claim generation \citep{jiang-etal-2025-large} or claims for specification drafting \citep{wang-etal-2024-patentformer}. As a result, current systems are often evaluated in settings that bypass the most challenging aspect of patent drafting: translating inventor-style disclosures into legally structured patent applications.

Recent work has begun to explore end-to-end patent generation, including draft-to-patent \citep{wang2024autopatent} and paper-to-patent settings \citep{knappich-etal-2025-pap2pat}. Although these approaches move closer to holistic drafting, their inputs are typically authored in formal or technical styles that differ substantially from authentic invention disclosures. Moreover, although patent figures play a central role in conveying technical structure and features in specific embodiments, especially in mechanical and system-oriented inventions, most existing datasets either omit figures or incorporate them only in limited ways \citep{jiang-etal-2025-large,jiang-etal-2025-enriching,jiang-etal-2025-patent, wang-etal-2024-patentformer, shea-yu-2025-autospec}. The role of multimodal inputs is widely underexplored.

In this work, we introduce Dis2Pat, a disclosure-to-patent dataset that targets a more realistic and practically relevant patent drafting scenario. Dis2Pat takes as input de-legalized, inventor-style disclosures accompanied by figures and requires models to generate a complete patent application, including both claims and a full specification. Because authentic invention disclosures are not publicly available due to confidentiality and legal constraints, we construct pseudo-disclosures from granted patents through a controlled LLM-based extraction process. This process preserves the underlying technical content but systematically removes clear invention feature structure, patent-specific legal language, and stylistic formats.

Building on Dis2Pat, we propose Patent-MAF, a multi-agent patent drafting framework designed to operate with locally deployable open-source models, motivated by the strong privacy and confidentiality requirements inherent in real-world patent drafting \citep{shea-yu-2025-autospec}. In many practical scenarios, invention disclosures cannot be shared with external APIs or proprietary services, which turns local model deployment into a necessity. Patent-MAF decomposes patent generation into coordinated roles: a manager agent analyzes and structures disclosures to route subtasks, specialized drafter agents generate claims and specifications, and a polisher agent finally refines cross-section consistency, legal style, and overall coherence.

The main contributions are as follows:

(1) We introduce Dis2Pat, the first dataset for disclosure-to-patent generation that incorporates inventor-style disclosures along with associated figures and patent texts.

(2) We propose a strong baseline named Patent-MAF, a multi-agent framework that operates a manager agent to structure the disclosure and coordinate specialized drafter agents, together with a polisher agent to refine final patent texts.

(3) We provide a systematic benchmark study on text-only, multimodal, open-source, and closed-source LLMs. We also demonstrate that Patent-MAF consistently outperforms all evaluated open-source models and achieves performance competitive with state-of-the-art closed-source models.

\section{Related Works}

Existing work on patent generation can be broadly categorized into three tasks: claim-centric generation, specification generation, and end-to-end patent-application drafting (Table~\ref{tab:patent-datasets}).

\noindent \textbf{Claim Drafting.}
Early studies primarily focused on generating patent claims from short inputs such as patent abstracts \citep{lee2020patentgenerate, lee2020controlling}. As language models recently became stronger and supported longer context lengths, subsequent work introduced large-scale datasets such as HUPD-DCG \citep{jiang-etal-2025-large} and EPD \citep{jiang-etal-2025-enriching} to investigate claim generation from full patent specifications. These studies highlighted the difficulty of faithfully recovering the inventive scope from long and highly technical inputs. Complementary to direct generation, Patent-CR \citep{jiang-etal-2025-patent} formulated claim revision as a post-editing task, which paired rejected application claims and granted claims to study legality-aware editing under realistic patent prosecution trajectories.

\noindent \textbf{Specification Drafting.}
Another line of research shifted attention to patent specification generation, which requires producing long-form, coherent, and technical text. PatentFormer \citep{wang-etal-2024-patentformer} explored specification generation conditioned on claims and figure descriptions. Subsequent work further expanded the range of input modalities: PatentVision \citep{yang2025patentvision} incorporated visual information from patent figures at scale, whereas PatentDesc \citep{shukla2025patentlmm} investigated the generation of specifications directly from figures alone. AutoSpec \citep{shea-yu-2025-autospec} examined a more constrained setting, which focused on the generation of specifications solely from claims.

\noindent \textbf{Patent Application Drafting.}
Beyond individual components, several recent datasets and tasks have targeted end-to-end patent drafting, where systems generate both claims and specifications from a single input. Draft2Patent \citep{wang2024autopatent} formulated a task that transforms patent drafts into complete patent applications, while Pap2Pat \citep{knappich-etal-2025-pap2pat} investigated the conversion of scientific articles into patent applications. Dis2Pat focuses on the disclosure-to-patent setting, in which the input consists of a natural-language patent disclosure written in a nonlegal, inventor-style form accompanied by drawings. This setting more closely aligns with real-world patent drafting workflows, where inventions are initially described informally and supported by illustrative drawings before being translated into features, legally structured claims, and specifications.

\section{The Dis2Pat Dataset}

\subsection{Construction}

\noindent \textbf{Source Corpus.}
Due to confidentiality and legal constraints, real-world invention disclosures are not publicly accessible. Thus, it is not possible to directly construct large-scale disclosure-to-patent datasets. To solve this limitation, we created a pseudo-disclosure dataset from publicly available granted patents with the help of LLMs. We extracted technical content, systematically removed patent-specific legal style, and blurred the feature-focused style of patent language.

In real-world patent drafting, invention disclosures typically include not only textual descriptions but also illustrative figures that convey core technical ideas. The PatentDesc dataset \citep{shukla2025patentlmm} filtered patents with high-quality drawings and was originally designed for figure-to-specification generation. We therefore adopt the patent publication numbers (the unique identifier assigned to a patent application when it is published) provided by PatentDesc as the source corpus to build our dataset.

\noindent \textbf{Filtering Criteria.}
We first extracted all publication numbers from the PatentDesc corpus and retained only granted patents. For each selected patent, we retrieved the title, abstract, claims, full specification, and figure image URLs from the Google Patents Public Dataset\footnote{\url{https://patents.google.com/}} to ensure consistent and up-to-date document components. To guarantee data completeness and multimodal consistency, we applied the following filtering criteria: (1) Document completeness: Each patent must contain all required components, including an abstract, a complete claim set, a full specification, and associated figures. (2) Figure constraints: Each patent must contain no more than ten figures, which excludes cases with excessively many figures that may introduce unnecessary visual complexity. After filtering, we randomly sampled 9,433 patents, which were split into 8,490 training and 943 test samples. We additionally verify patent-family metadata to ensure that no patent family appears in both the training and test sets. Because Dis2Pat evaluates conditional drafting from a given disclosure rather than temporal generalization, we do not impose a temporal split.

\noindent \textbf{Pseudo-Disclosure Generation.}
For each selected patent, we generated a corresponding pseudo-disclosure with GPT-5-mini\footnote{\url{https://platform.openai.com/docs/models/gpt-5-mini}}. The model was prompted to rewrite the original patent text into a clear, human-readable invention disclosure using plain, nonlegal English, but to preserve factual and technical content. The full prompt is provided in Appendix~\ref{apx:dataset}.

Each pseudo-disclosure is organized into seven invention-centric components: (1) Title of the Invention, (2) Why This Invention Is Needed, (3) What the Core Idea Is, (4) How It Works, (5) What Makes It Different, (6) What Benefits It Provides, and (7) Optional Variants. This decomposition reflects how inventors typically communicate inventions prior to formal patent drafting. Overall, these components provides sufficient technical coverage for downstream patent drafting but avoids direct reuse of patent claim language. Dis2Pat deliberately assumes that the disclosure contains sufficient technical information for drafting and isolates the subsequent drafting stage; eliciting missing information from incomplete inventor notes is treated as a distinct upstream problem. Information-complete does not imply patent-structured: the disclosures omit explicit claim scope, claim hierarchies, and specification organization.

\input{tabs/tab_datastats}

\subsection{Statistics and Quality Analysis}

Table~\ref{tab:dataset_statistics} summarizes the corpus statistics and quality analysis of our dataset. On average, each original patent contains 11,207 tokens, including 1,594 tokens in claims and 9,477 tokens in the specification, together with 6.7 figures and 19.7 claims per document. In contrast, the corresponding pseudo-disclosures are substantially more concise, with an average length of 1,196 tokens. This reduction reflects the removal of patent-specific legal structure while preserving the core technical content. The pronounced decrease in both token count and technical term density (from 0.05 to 0.001) further indicates that the pseudo-disclosures effectively eliminate legalistic and repetitive patent language.

To assess the faithfulness and quality of the generated pseudo-disclosures, we randomly sampled 100 examples for human evaluation. Patent professionals evaluated each disclosure along four dimensions: (1) Hallucination: degree to which the disclosure contains statements not supported by the patent. (2) Missing Details: degree to which the disclosure omits essential technical information present in the patent. (3) Contradiction: degree to which the disclosure contradicts the patent. (4) De-legalization: how effectively the disclosure removes legalistic patent language but keeps the technical meaning correct. 
All evaluation scores are reported on a ten-point scale, where higher values indicate better quality. As summarized in Table~\ref{tab:dataset_statistics}, the pseudo-disclosures achieve consistently high scores across all metrics, including Hallucination (9.8), Details (9.8), Contradiction (10.0), and De-legalization (9.7). The results indicate that the LLM-based rewriting process preserves factual and technical content with minimal hallucination or inconsistency, and successfully removes patent-specific legal language. To further quantify surface-form divergence from the source patents, we measure lexical overlap. The average ROUGE-L F1 is 0.08 (maximum 0.11), while the average 4-gram overlap is 0.006 (maximum 0.013), which indicates that the pseudo-disclosures are not lexical copies of their source patents.

To further examine the stability of these evaluations, we measure the effect size between evaluation subsets using Cohen’s d. Specifically, the 100 validated examples are randomly divided into two equally sized groups, and the resulting effect size is approximately 0.1, which corresponds to a negligible effect according to standard benchmarks. This small effect size suggests that score variations across samples are minimal, which provides additional evidence for the reliability and robustness of the dataset. Thus, Dis2Pat provides a controlled and scalable proxy for the patent-drafting stage.

\begin{figure*}[!t]
    \centering   
    \includegraphics[width=\textwidth]{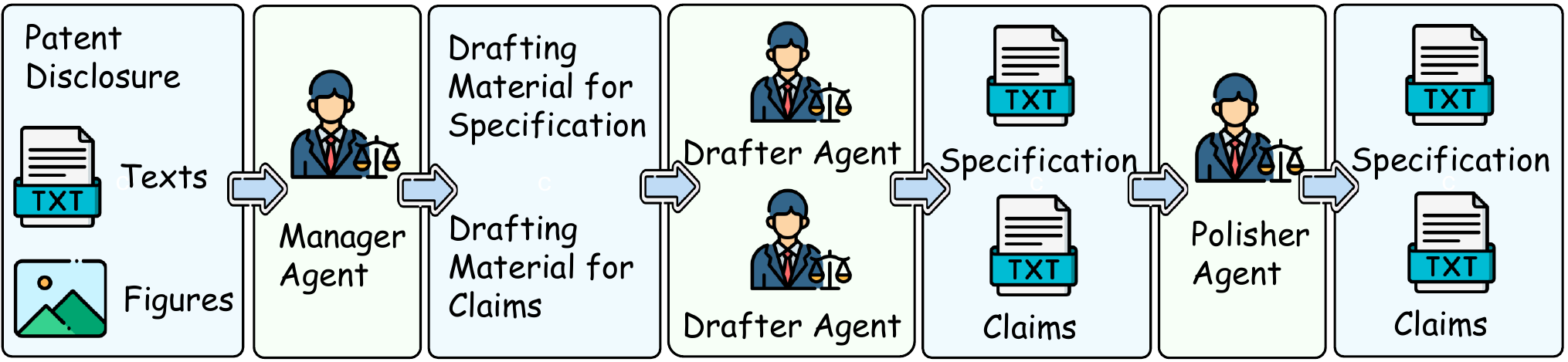}   
    \caption{Overview of the Patent-MAF method. The manager agent structures the disclosure and coordinates specialized drafter agents, followed by a polisher agent to refine the final patent text.}
    \label{fig:overview}
\end{figure*}

\section{Patent-MAF}

We propose a multi-agent baseline for generating patent claims and specifications from invention disclosures, namely \textit{Patent-MAF}. Figure~\ref{fig:overview} illustrates the overall workflow of Patent-MAF. The framework first commands a manager agent to structure the disclosure and coordinate specialized drafting agents, followed by a polisher agent that refines the generated patent text into a coherent and legally consistent application.

\noindent \textbf{Problem Formulation.}
Let $D = (T, F)$ denote a patent disclosure, where $T$ is a de-legalized textual description of the invention and $F$ is a set of associated figures. The goal is to generate a patent draft $P = (C, S)$, consisting of a claim set $C$ and a specification $S$.

\noindent \textbf{Manager Agent.}
The \textit{manager agent} serves as the central coordinator of the framework. Its role is to transform the raw disclosure $D$ into structured drafting materials $M = (M_C, M_S)$ that are respectively tailored to claim drafting and specification drafting. Specifically, the Manager Agent produces two structured artifacts: (1) claim drafting material $M_C$, which emphasizes legally essential features and scope-defining elements; and (2) specification drafting material $M_S$, which emphasizes technical background, detailed explanations, embodiments, and references to figures. The manager agent is implemented using prompt-based reasoning with explicit output schemata to ensure transparency and consistency.

\noindent \textbf{Drafter Agents.}
Patent-MAF uses specialized \textit{drafter agents} to generate different components of the patent, each operating independently while being grounded in the manager agent’s outputs.
The \textit{claim drafter agent} generates the complete set of patent claims $C$ from $M_C$, including at least one independent claim and multiple dependent claims. Given the highly formalized and legally constrained nature of claim drafting, this agent can optionally be fine-tuned to improve structural correctness and legal robustness.
The \textit{specification drafter agent} generates the patent specification $S$ from $M_S$, including the background, summary, and detailed embodiments. This agent prioritizes technical clarity and completeness and adheres to conventional patent-style narrative structure.

\noindent \textbf{Polisher Agent.}
The independently generated claims and specification are subsequently passed to the \textit{polisher agent}, whose role is global refinement rather than content creation. Specifically, the polisher agent performs terminology unification across claims and specification, consistency checking to eliminate mismatches between claim language and descriptive text, and stylistic normalization to ensure a professional patent writing tone. Importantly, the polisher agent does not introduce new technical content, but only revises existing text to improve coherence, consistency, and correctness.

\noindent \textbf{Open-Source Model Requirement. }
A core design requirement of Patent-MAF is that all agents operate on locally deployable, open-source language models. This requirement is motivated by the privacy and confidentiality constraints inherent in real-world patent drafting, where invention disclosures often contain sensitive technical information and cannot be shared with external APIs or proprietary services. As a result, frameworks that rely on closed-source or cloud-hosted models are impractical in many realistic deployment scenarios.

Patent-MAF differentiates itself from prior agentic patent systems in: (1) privacy-aware local deployment: all agents run on open-source models without any API calls to proprietary services as an unnegotiable constraint in real IP practice; (2) domain-grounded role decomposition: the manager, drafter, and polisher roles directly mirror the division of labor in professional patent prosecution rather than being generic task decomposition; (3) multimodal grounding: the specification drafter is vision-enabled to resolve spatial and structural ambiguities in figures.

\section{Experiments}

\subsection{Models}

We evaluate all models under the same decoding configuration to ensure a fair comparison. Experimental details are provided in Appendix~\ref{apx:models}.

\noindent \textbf{Text-Only Inputs.}
We compare against a diverse set of strong LLMs, including Qwen3-32B, Qwen3-VL-32B, Qwen3-VL-72B \citep{yang2025qwen3}, LLaMA-3.3-70B, LLaMA-3.2-90B-Vision \citep{dubey2024llama}, GPT-4o, and GPT-5 \citep{openai_2025}. These models generate patent claims and specifications directly from disclosure text without access to visual information. Due to the extreme length of patent specifications, we do not fine-tune models for specification generation. In contrast, patent claims are substantially shorter and subject to strict stylistic and structural constraints, which makes them suitable for parameter-efficient fine-tuning. Accordingly, we apply LoRA fine-tuning \citep{hu2021lora} to Qwen3-32B for claim generation to improve legal consistency and structural normalization.

\noindent \textbf{Text and Image Inputs.}
To assess the impact of visual information, we additionally evaluate multimodal baselines that take both disclosure text and associated figures as input, including Qwen3-VL-32B, Qwen3-VL-72B, LLaMA-3.2-90B-Vision, GPT-4o, and GPT-5. This setting reflects recent multimodal approaches to patent drafting that incorporate figures during generation.

\noindent \textbf{Ours.}
Patent-MAF adopts a multi-agent drafting framework in which different agents are instantiated with task-specific models. We implemented both the manager agent, responsible for disclosure structuring and task routing, and the polisher agent, responsible for global consistency and style refinement, based on Qwen3-32B with prompting. For content generation, we use two specialized drafter agents: the specification drafter is based on Qwen3-VL-32B to process both textual disclosures and associated figures, whereas the claim drafter is based on Qwen3-32B with fine-tuning to better capture the strict stylistic conventions of patent claims. The Qwen3-32B-FT baseline uses exactly the same fine-tuned checkpoint as the Patent-MAF claim drafter; the difference is direct generation versus the full agentic workflow. In preliminary experiments, we observe that visual inputs provide limited benefit for claim drafting, as claims primarily encode an abstract legal scope rather than a concrete visual structure. We therefore decided for text-only input for claim generation. 

We report results for the full Patent-MAF system as well as the following ablation variants: \textit{(i)} w/o manager, which removes disclosure structuring and task routing; \textit{(ii)} w/o visual input for specification generation or w/o fine-tuning for claim generation; and \textit{(iii)} w/o polisher, which omits the final consistency and style refinement stage. 

\input{tabs/tab_spec}

\subsection{Evaluation Metrics}

We evaluate generated patent specifications and claims with a combination of automatic metrics, LLM-as-a-judge quality assessment, and expert human evaluation.

\noindent \textbf{Text Overlap Metrics.}
We report BLEU \cite{papineni2002bleu}, ROUGE-1 (R-1), and ROUGE-L (R-L) \citep{lin2004rouge} for both patent specifications and claims to measure surface-level lexical overlap with the corresponding gold references.

\noindent \textbf{Semantic Similarity Metrics.}
To assess semantic alignment beyond word overlap, we additionally report BERTScore \cite{zhang2019bertscore} and BERT-for-Patent embedding similarity \citep{chikkamath2022patent} for both specifications and claims.

\noindent \textbf{LLM-Based Content Quality Evaluation.}
We further assess content quality using LLM-as-a-judge methods \cite{li-etal-2025-generation}. For both specifications and claims, we design task-specific evaluation prompts that compare generated outputs against gold references and assign scores on a 0--100 scale where higher scores indicate better quality. Detailed prompt is reported in Appendix \ref{apx:evaldetails}. 

For specifications, the evaluation focuses on: (1) completeness (whether the technical background, problem formulation, core solution, implementation details, and embodiments are sufficiently covered); (2) logical clarity (structural coherence and technical flow); and
(3) legal quality (adherence to patent drafting norms). For claims, the evaluation emphasizes:
(1) technical scope (whether the candidate claims capture the core inventive concept of the gold claims); (2) claim legality (compliance with professional claim-drafting standards); and
(3) logical clarity (technical clarity and unambiguous terminology).

\noindent \textbf{Expert Human Evaluation.}
To complement automatic evaluation, we conduct expert human assessment on 60 randomly sampled test cases. To reduce annotation cost and still achieve a high evaluation reliability, we adopt a pairwise comparison protocol rather than absolute scoring. For each sample, patent experts are presented with outputs from two systems and asked to determine which is superior. The comparison focuses on the first independent claim and selective specification segments that describe the invention. Experts judge each pair according to four criteria commonly used in patent examination: completeness, clarity, legality, and consistency between claims and specification. Appendix \ref{apx:humaneval} introduces human evaluation details. We further report agreement between automated and expert judgments in Appendix \ref{apx:evaldetails} to contextualize the reliability of automatic evaluation.

\section{Results and Analysis}

\subsection{Results on Specification Generation}

Table~\ref{tab:spec} reports model performance on patent specification generation. We analyze the results from the following perspectives.

\noindent \textbf{Effect of Visual Inputs.}
The comparison of text-only and text+image settings demonstrates that the incorporation of figures leads to consistent, slight improvements across models with vision encoders. For example, Qwen3-VL-32B improves the overall content quality score from 81.0 (text-only) to 81.3 (text+image), while Qwen3-VL-72B improves from 81.1 to 81.8. Similar trends are observed for semantic similarity metrics: for GPT-5, the BERT-for-Patent similarity increases from 95.7 to 96.4 when visual inputs are enabled. A possible explanation is that vision provides grounding for spatial and structural descriptions. When checking specific cases, we find that text-only models mis-describe component relationships, but the vision-enabled drafter correctly resolves them by referencing figures. These results suggest that figures provide complementary information for specification drafting, which particularly benefits content completeness.

\noindent \textbf{Advantage of Patent-MAF.}
Patent-MAF achieves the strongest overall performance among all evaluated open-source systems and exhibits performance comparable to or even better than state-of-the-art closed-source models. In particular, Patent-MAF reaches an overall content quality score of 85.4, which outperforms strong open-source baselines such as Qwen3-VL-72B (81.8) and closed-source GPT-5 (85.1). Patent-MAF also achieves competitive semantic similarity, with a BERTScore of 82.3 and a BERT-for-Patent score of 96.1, compared to GPT-5’s 81.8 and 96.4, respectively. The results indicate that structured multi-agent orchestration can improve specification generation quality and narrow the gap between small-sized open-source models and state-of-the-art closed-source systems.

\input{tabs/tab_claim}

\noindent \textbf{Ablation Study.}
The ablation results further highlight the contribution of individual components in Patent-MAF. Removing the manager agent reduces the overall score from 85.4 to 83.4, accompanied by declines in completeness (from 85.6 to 85.0) and clarity (from 88.5 to 86.9), which underscores the importance of structured information organization. Disabling visual inputs similarly lowers the overall score from 85.4 to 83.4 and reduces BERT-for-Patent similarity from 96.1 to 93.6. Thus, figures tend to improve semantic alignment. Removing the polisher agent primarily affects global coherence and stylistic quality, illustrated by the decrease in the style score from 82.0 to 80.9. 

\begin{figure*}[!t]
    \centering   
    \includegraphics[width=\textwidth]{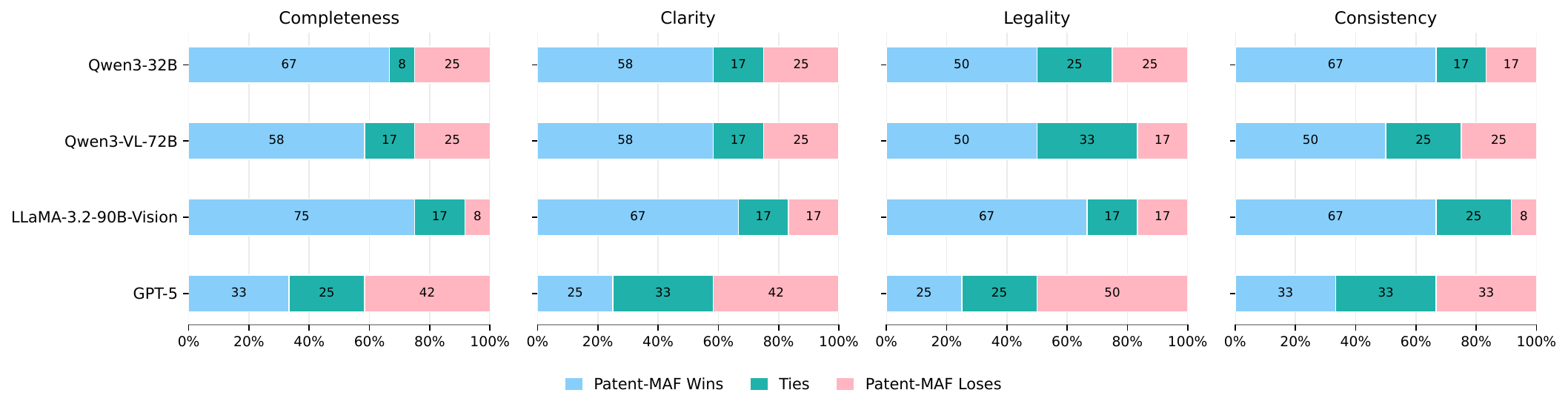}   
    \caption{Pairwise human evaluation results that compare Patent-MAF against four strong baselines. }
    \label{fig:humaneval}
\end{figure*}

\subsection{Results on Claim Generation}

Table~\ref{tab:claim} reports the performance of different models on patent claim generation. 

\noindent \textbf{Effect of Visual Inputs.}
Overall, incorporating visual inputs does not bring performance gains in claim generation. For example, Qwen3-VL-32B shows a slight decrease in overall score from 78.9 (text-only) to 78.6 (text+image), while Qwen3-VL-72B drops marginally from 80.7 to 79.8. A similar trend is observed for GPT-5, whose overall score decreases from 91.1 (text-only) to 90.7 (text+image) when visual inputs are enabled. This trend is also extended to text overlap and semantic similarity metrics. The results suggest that, unlike specification drafting, patent claim generation relies primarily on precise textual modeling of inventive scope and legal structure, and figures provide no obvious benefit.

\noindent \textbf{Advantage of Patent-MAF.}
Patent-MAF achieves the strongest overall performance among all evaluated open-source systems, with an overall quality score of 86.7. Although the closed-source GPT-5 reaches the highest overall quality score of 91.1, Patent-MAF achieves higher semantic similarity with a BERTScore of 88.3 compared to GPT-5’s 83.8, and a BERT-for-Patent score of 96.4 compared to GPT-5’s 95.9. The results indicate that the proposed multi-agent framework substantially narrows the performance gap and demonstrates strong alignment with domain-specific claim semantics.

\noindent \textbf{Ablation Study.}
The ablation results further illustrate the contribution of individual components in Patent-MAF. Removing the manager agent reduces the overall score from 86.7 to 84.2, which highlights the importance of structured disclosure organization for defining claim scope. Removing fine-tuning lowers the overall score by 5.1 points (86.7 to 81.6), which confirms the importance of task-specific training for claim drafting. Removing the polisher causes the largest degradation, a 6.6-point drop (86.7 to 80.1), which highlights its role in legal consistency and stylistic correctness.

\subsection{Overall Results}

The results reveal distinct behaviors for claim generation and specification drafting. Specification drafting benefits more substantially from multimodal inputs and global consistency refinement, which leads to noticeable improvements in completeness and clarity. Claim generation, in contrast, is more sensitive to precise textual modeling and adherence to legal conventions, where task-specific fine-tuning plays a more central role.

Figure~\ref{fig:humaneval} presents the results of pairwise expert evaluations, where annotators compare Patent-MAF with four strong baselines (Qwen3-32B, Qwen3-VL-72B, LLaMA-3.2-90B-Vision, and GPT-5) using the first independent claim and specification segments that describe the core invention. Against all open-source baselines, Patent-MAF is preferred in the majority of cases across all four criteria (completeness, clarity, legality, and consistency). The results indicate that the proposed agent framework is effective in patent drafting beyond single-shot generation.

Comparisons with GPT-5 show a different pattern. GPT-5 is preferred more often on completeness, clarity, and legality, while the two systems are evenly matched on consistency. At the same time, the substantial proportion of ties indicates that Patent-MAF remains competitive with GPT-5, particularly in maintaining cross-section consistency.Thus, explicit task decomposition and refinement appear particularly beneficial for ensuring comprehensive coverage and maintaining coherence between claims and specifications, even when compared with a state-of-the-art closed-source model. More results, including qualitative analysis, are introduced in Appendix \ref{moreresults}.

\section{Conclusion} 
We introduce Dis2Pat, the first dataset designed to model realistic disclosure-to-patent drafting. We propose Patent-MAF as a strong baseline, a multi-agent framework optimized for locally deployable open-source models. By eliminating reliance on external APIs, Patent-MAF safeguards sensitive invention disclosures. The framework decomposes the drafting process into coordinated roles: a manager agent analyzes the disclosure and routes subtasks; specialized drafter agents generate claims and specifications; and a polisher agent refines the output for cross-section consistency and legal coherence. Overall, the results establish Dis2Pat as a challenging benchmark for disclosure-to-patent drafting and show that our multi-agent framework is a strong locally deployable baseline.

\section*{Limitations}

Despite promising results, this work has several limitations that suggest directions for future research. Due to confidentiality and legal constraints, Dis2Pat relies on pseudo-disclosures generated from granted patents via controlled LLM-based extraction and rewriting. Although we carefully constrain the rewriting process and validate disclosure quality through both automatic and expert evaluations, these pseudo-disclosures may not fully capture the diversity, incompleteness, or ambiguity present in real-world invention disclosures submitted by inventors. Future work could explore partnerships with industry or law firms to study anonymized or partially synthetic disclosures. In addition, Patent-MAF is evaluated primarily on English-language patents. Adapting the framework to multilingual or jurisdiction-specific patent drafting standards (e.g., EPO or CNIPA) is a natural extension. We do not perform hyperparameter tuning during inference; instead, we keep the hyperparameters fixed across all experiments to ensure fair comparisons.

\section*{Ethics Statement}
Llama-3 is under \textit{META LLAMA 3 COMMUNITY LICENSE AGREEMENT}. Qwen-3 is under \textit{Apache license 2.0}. GPT-5 and GPT-5-mini are under a commercial license provided by OpenAI, and we access them through its API. The use of existing artifacts is consistent with their intended use. The data does not include potential personal information or offensive content, and no ethics review board was involved. Our proposed dataset will be under the \textit{CC-BY-SA-4.0} license.

\section*{Acknowledgments}
This research was supported in part by Lambda, Inc.

\bibliography{anthology,custom}

\appendix

\input{tabs/tab_datacreate}

\section{Patent Background}
\label{apx:background}

Patent documents are a primary mechanism for safeguarding intellectual property (IP) and for communicating technical innovations. They offer a legally enforceable representation of an invention by specifying the scope of exclusive rights conferred to the patent owner. After publication or grant, patents become publicly available and constitute a key source of technical disclosure for both legal practice and scientific research.

Despite jurisdictional differences in formatting, patent documents generally follow a standardized structure. Core elements typically include a title, bibliographic information, classification codes, prior-art citations, an abstract, illustrative figures, a detailed description (specification), and a set of claims. Among these components, the claims carry legal authority by precisely delineating the extent of patent protection, whereas the specification provides supporting technical detail, embodiments, and contextual explanations that ground and interpret the claims.

From the perspective of natural language processing (NLP), patents present distinctive challenges. Prior work \citep{jiang2025natural} highlights that patent-related tasks require models to reason over extremely long contexts that span multiple sections and visual elements, to accurately interpret highly specialized and technical language, and to maintain strict precision and internal consistency imposed by legal requirements. These factors render patent drafting substantially more complex than typical text generation tasks, thereby motivating the development of specialized modeling and evaluation approaches. Recent research on patent NLP or LLMs include classification \citep{jiang2026rhc, yoo2025self}, novelty prediction \citep{knappich2026novel}, generation \citep{jiang2026reasoning, yoo2026pointer}, quality assessment \citep{song2026pquasar}, evaluation \citep{zuo-etal-2024-patenteval,yoo2025patentscore, yoo-etal-2026-patentmind, jiang2025patclaimeval,yoo2026patdeval,yoo2026adaptive}.

\section{Dataset Details and Comparison with Prior Works}
\label{apx:dataset}

This section provides additional details on the construction of the Dis2Pat dataset. Table~\ref{tab:disclosure-prompt} introduces the prompt to create the pseudo-disclosure.

Dis2Pat differs from Draft2Patent in task formulation, input representation, and methodological focus. Draft2Patent uses answers to five predefined patent-oriented questions, including explicit protection intent, whereas Dis2Pat uses de-legalized, inventor-oriented disclosures that preserve technical content but omit explicit claim scope. Draft2Patent uses text-only inputs with textual figure descriptions, while Dis2Pat retains the original figures as multimodal inputs. Methodologically, Draft2Patent primarily focuses on hierarchical expansion into a long patent document, whereas Patent-MAF additionally structures disclosure information into claim- and specification-specific materials and performs cross-section consistency refinement. We do not include Draft2Patent as a direct baseline because the input formats are not directly compatible and its implementation is not publicly available.

\input{tabs/tab_cross}

\section{Models and Implementation Details}
\label{apx:models}

All fine-tuning and inference experiments are conducted on NVIDIA A100 GPUs. The total compute time across all experiments is approximately 700 GPU hours. We adopt parameter-efficient fine-tuning using LoRA. Specifically, we set the LoRA rank to 16, LoRA alpha to 8, and apply a dropout rate of 0.05. The batch size is set to two due to long input sequences, with a learning rate of $5\times10^{-5}$ and weight decay of 0.1. Models are trained for up to three epochs. During inference, we use a temperature of 0.3 to reduce randomness. The maximum number of generated tokens is set to 16,384. A unified prompt format is used across all models to ensure consistency and isolate the effects of model architecture and training strategies.

\section{Evaluation Details}
\label{apx:evaldetails}

Previous work has shown that LLM-as-a-judge evaluators can achieve better human alignment \citep{liu-etal-2023-g}. Thus, we use Deepseek-V3 \citep{liu2024deepseek} with Chain-of-Thought (CoT) \citep{wei2022chain} prompting to evaluate the generated specification and patent claims. This section provides additional details on the automatic evaluation metrics used in our experiments. Tables~\ref{tab:spec-eval-prompt} and~\ref{tab:claim-eval-prompt} report the full prompt for specification and claim evaluation, respectively. Furthermore, we conduct and report correlation analysis comparing LLM scoring preferences against expert pairwise judgments. Agreement rates are: Completeness 82\%, Consistency 76\%, Clarity 68\%, and Legality 62\%, indicating substantial alignment across all four dimensions. As a judge-robustness check, we additionally repeat the evaluation with GLM-5.2 using the same prompts and settings. Its pairwise preferences agree with DeepSeek-V3 on approximately 85\% of specification comparisons and 82\% of claim comparisons. 

\input{tabs/tab_speceval}
\input{tabs/tab_claimeval}

\section{Human Evaluation Details}
\label{apx:humaneval}

We invite licensed patent attorneys to volunteer for human evaluations. These professionals are provided with reference claims and candidate claims for assessment. They are informed about the intended use of the evaluation results. Table \ref{tab:human-eval-instructions} introduces detailed instructions. 

Due to the high cost of expert review in the patent domain, we conducted human evaluation with two annotators: one licensed patent attorney and one experienced patent practitioner-in-training. This design prioritizes domain expertise and legally informed judgment over large-scale crowd annotation. For each evaluation dimension, we measure inter-annotator agreement (IAA) using weighted Cohen's kappa ($\kappa$). The agreement scores are $\kappa = 0.79$ for Completeness, $\kappa = 0.73$ for Consistency, $\kappa = 0.71$ for Clarity, and $\kappa = 0.65$ for Legality, which indicates substantial agreement overall. This level of agreement is consistent with the greater subjectivity and jurisdiction-sensitive nature of legal drafting judgments. For the final pairwise decisions reported in our human evaluation, we use the licensed patent attorney's judgments as the primary expert decision, while the second annotator's judgments are used to assess annotation reliability through IAA. To improve transparency, we provide the full evaluation rubric in Table~\ref{tab:human-eval-instructions}. 

\input{tabs/tab_humaneval}

\section{More Results}
\label{moreresults}

\subsection{Cross-model Sensitivity Test}
We extend experiments by instantiating Patent-MAF with Llama-3.2-90B-Vision in Table \ref{tab:cross} and observed similar performance trends, which supports that the framework generalizes across open-source backbones.

\subsection{Statistical Analysis}
Regarding statistical analysis, we conduct paired t-tests on all automated metrics comparing Patent-MAF against baselines. All reported improvements are statistically significant (p < 0.05).

\subsection{Computational Overhead}
Regarding computational overhead, Patent-MAF incurs approximately 2.5x token usage relative to a single-shot baseline, with proportionally increased latency due to the Manager and Polisher agents' intermediate reasoning steps.

\subsection{Qualitative Analysis}

We identify three primary error patterns. First, Manager routing omissions: when the Manager fails to extract a key technical feature from the disclosure, downstream Drafter agents generate incomplete claims or specifications, and the Polisher--which performs consistency rather than completeness checking--cannot recover the missing information. Second, cross-section terminology inconsistency: because Claim and Specification Drafters operate independently, they occasionally use divergent terms for the same objects; the Polisher successfully resolves most such cases, but residual mismatches persist when the inconsistency spans long contexts. Third, spatial description errors in text-only generation: when figure information is absent, the Specification Drafter sometimes misidentifies component; the vision-enabled Drafter resolves these by grounding descriptions in the figures, which provides a concrete justification for multimodal integration despite its marginal aggregate gains. 

\end{document}

%% file: tabs/tab_relatedwork.tex
\begin{table*}[!t]
\centering
\footnotesize

\resizebox{\textwidth}{!}{
\begin{tabular}{l |c |l |l |l}
\toprule
\textbf{Dataset / Work} & \multicolumn{1}{c|}{\textbf{Size}} & \multicolumn{1}{c|}{\textbf{Task}} & \multicolumn{1}{c|}{\textbf{Input}} & \multicolumn{1}{c}{\textbf{Output}} \\
\midrule
HUPD-DCG \citep{jiang-etal-2025-large}& $\sim$10K & Claim generation & Specification & Claims \\
EPD \citep{jiang-etal-2025-enriching} & $\sim$107K & Claim generation & Specification & Claims \\
Patent-CR \citep{jiang-etal-2025-patent} &  $\sim$20K & Claim revision & Rejected application claims & Granted claims \\
PatentFormer \citep{wang-etal-2024-patentformer}&  $\sim$13K & Specification generation & Claims + figure descriptions & Specification \\
PatentVision \cite{yang2025patentvision} & $\sim$230K & Specification generation & Claims + figure & Specification \\
AutoSpec \cite{shea-yu-2025-autospec} & $\sim$1.3K & Specification generation & Claims & Specification \\
PatentDesc \citep{shukla2025patentlmm} & $\sim$355K & Specification generation & Figures & Specification \\ 
Draft2Patent \cite{wang2024autopatent} &  $\sim$2K & Draft-to-patent & Patent draft & Patent application \\
Pap2Pat \citep{knappich-etal-2025-pap2pat} &  $\sim$1.8K & Paper-to-patent & Scientific paper & Patent application \\
\rowcolor{gray!30}
Dis2Pat (Ours) & $\sim$9.4K & Disclosure-to-patent & Patent disclosure & Patent application  \\
\bottomrule
\end{tabular}
}
\caption{Comparison of related datasets and works for LLM-based patent drafting.}
\label{tab:patent-datasets}
\end{table*}

%% file: tabs/tab_datastats.tex
\begin{table}[!t]
\centering
\footnotesize
\resizebox{.48\textwidth}{!}{
\begin{tabular}{lcc}
\toprule
\textbf{Metric} & \textbf{Original Patent} & \textbf{Disclosure} \\
\midrule
\multicolumn{3}{l}{\textbf{Corpus Statistics}} \\
\# Tokens & 11207.4  & 1196.2 \\
\# Title Tokens & 8.7 & -- \\
\# Abstract Tokens & 128.1 & -- \\
\# Claim Tokens & 1594.1 & --\\
\# Specification Tokens & 9476.5 & --\\
\# Figures & 6.7 & --  \\
\# Claims & 19.7 & -- \\
Term Density & 0.05 & 0.001 \\
\midrule
\multicolumn{3}{l}{\textbf{Quality and Faithfulness}} \\
Hallucination Score & -- & 9.8 \\
Details Score & -- & 9.8 \\
Contradiction Score  & -- & 10.0 \\
De-legalization Score & -- & 9.7 \\
\bottomrule
\end{tabular}
}
\caption{Statistics and quality of the Dis2Pat dataset.}
\label{tab:dataset_statistics}
\end{table}

%% file: tabs/tab_spec.tex
\begin{table*}[!t]
\centering
\footnotesize
\resizebox{.99\textwidth}{!}{
\begin{tabular}{l|ccc|cc|cccc}
\toprule
\multirow{2.7}{*}{\textbf{Model / Setting}} & \multicolumn{3}{c|}{\textbf{Text Overlap}} & \multicolumn{2}{c|}{\textbf{Semantic Similarity}} & \multicolumn{4}{c}{\textbf{Content Quality}} \\
\cmidrule(lr){2-4}\cmidrule(lr){5-6}\cmidrule(lr){7-10}
 & \textbf{BLEU} & \textbf{R-1} & \textbf{R-L} & \textbf{BERTScore} & \textbf{BERT-for-Patent} & \textbf{Completeness} & \textbf{Clarity} &  \textbf{Style} & \textbf{Overall} \\
\midrule
\multicolumn{3}{l}{\textbf{Text-Only}} \\
Qwen3-32B 
& 0.8 & 28.8 & 12.1 & 81.5 & 94.4 & 82.8 & 81.0 & 72.9 & 78.9 \\
Qwen3-VL-32B 
& 0.7 & 30.3 & 12.3 & 82.0 & 94.6 & 83.3 & 85.9 & 73.7 & 81.0 \\
Qwen3-VL-72B 
& 1.6 & 32.2 & 12.9 & 81.6 & 94.5 & 84.2 & 83.5 & 75.7 & 81.1 \\
Llama-3.3-70B
& 0.3 & 26.1 & 11.3 & 81.6 & 94.6 & 72.0 & 74.7 & 59.6 & 68.8 \\
Llama-3.2-90B-Vision 
& 0.5 & 27.3 & 11.8 & 81.5 & 94.5 & 68.7 & 73.6 & 58.1 & 66.8 \\
GPT-4o 
& 0.1 & 12.3 & 6.5 & 81.7 & 87.2 & 72.4 & 77.0 & 64.1 & 71.2 \\
GPT-5 
& 1.5 & 31.1 & 11.9 & 81.8 & 95.7 & 84.2 & 88.4 & \textbf{82.4} & 85.0 \\

\midrule
\multicolumn{3}{l}{\textbf{Text+Image}} \\
Qwen3-VL-32B
& 0.8 & 30.5 & 12.5 & 82.0 & 94.7 & 83.6 & 86.1 & 74.3 & 81.3 \\
Qwen3-VL-72B
& 1.8 & 32.4 & 12.9 & 81.6 & 94.7 & 84.8 & 83.6 & 77.1 & 81.8 \\
Llama-3.2-90B-Vision 
& 0.6 & 27.7 & 12.1 & 81.8 & 94.6 & 68.9 & 73.9 & 58.6 & 67.1 \\
GPT-4o
& 0.4 & 13.4 & 6.8 & 81.9 & 89.2 & 75.5 & 79.6 & 65.9 & 73.7 \\
GPT-5
& \textbf{2.3} & \textbf{34.8} & \textbf{13.9} & 81.8 & \textbf{96.4} & 84.7 & 88.2 & 82.3 & 85.1 \\
\midrule
\multicolumn{3}{l}{\textbf{Ours (Based on Qwen3-VL-32B)}} \\
Patent-MAF
& 2.1 & 32.5 & 13.8 & \textbf{82.3} & 96.1 & \textbf{85.6} & \textbf{88.5} & 82.0 & \textbf{85.4} \\
\quad w/o Manager 
& 0.8 & 27.7 & 11.5 & 81.6 & 94.1 & 85.0 & 86.9 & 78.5 & 83.4 \\
\quad w/o Vision 
& 0.1 & 22.8 & 10.2 & 81.6 & 93.6 & 85.4 & 86.8 & 77.9 & 83.4 \\
\quad w/o Polisher 
& 1.2 & 31.8 & 12.9 & 81.8 & 95.1 & 85.3 & 87.1 & 80.9 & 84.4 \\
\bottomrule
\end{tabular}
}
\caption{Model performance on patent specification. The highest score in each column is marked in \textbf{bold}.  }
\label{tab:spec}
\end{table*}

%% file: tabs/tab_claim.tex
\begin{table*}[!t]
\centering
\footnotesize

\resizebox{.98\textwidth}{!}{
\begin{tabular}{l|ccc|cc|cccc}
\toprule
\multirow{2.7}{*}{\textbf{Model / Setting}} & \multicolumn{3}{c|}{\textbf{Text Overlap}} & \multicolumn{2}{c|}{\textbf{Semantic Similarity}} & \multicolumn{4}{c}{\textbf{Content Quality}} \\
\cmidrule(lr){2-4}\cmidrule(lr){5-6}\cmidrule(lr){7-10}
 & \textbf{BLEU} & \textbf{R-1} & \textbf{R-L} & \textbf{BERTScore} & \textbf{BERT-for-Patent} & \textbf{Scope} & \textbf{Clarity} &  \textbf{Legality} & \textbf{Overall} \\
\midrule
\multicolumn{10}{l}{\textbf{Text-Only}} \\
Qwen3-32B 
& 13.5 & 48.0 & 25.3 & 83.2 & 94.5 & 82.8 & 81.2 & 78.6 & 80.9 \\
Qwen3-32B-FT 
& 19.1 & 54.3 & 30.4 & 87.3 & 96.1 & 83.6 & 85.2 & 81.2 & 84.4 \\
Qwen3-VL-32B 
& 10.8 & 48.6 & 24.1 & 82.3 & 93.9 & 82.7 & 79.5 & 74.6 & 78.9 \\
Qwen3-VL-72B 
& 13.5 & 51.3 & 26.3 & 84.3 & 94.5 & 83.5 & 81.1 & 77.4 & 80.7 \\
Llama-3.3-70B
& 9.5 & 43.3 & 22.8 & 81.9 & 93.3 & 76.3 & 77.5 & 77.8 & 77.2 \\
Llama-3.2-90B-Vision 
& 9.8 & 44.6 & 24.5 & 83.5 & 93.8 & 78.5 & 78.8 & 79.3 & 78.9 \\
GPT-4o 
& 1.0 & 24.9 & 15.3 & 82.4 & 86.6 & 67.8 & 72.7 & 71.6 & 70.7 \\
GPT-5 
& 19.0 & 53.8 & 26.8 & 83.8 & 95.9 & \textbf{85.8} & \textbf{95.1} & \textbf{92.3} & \textbf{91.1} \\

\midrule
\multicolumn{10}{l}{\textbf{Text+Image}} \\
Qwen3-VL-32B
& 10.9 & 48.6 & 24.1 & 82.3 & 93.9 & 81.4 & 78.3 & 76.2 & 78.6 \\
Qwen3-VL-72B
& 12.4 & 50.1 & 25.2 & 83.7 & 94.2 & 82.9 & 80.0 & 76.5 & 79.8 \\
Llama-3.2-90B-Vision
& 9.1 & 43.3 & 23.2 & 82.3 & 93.5 & 78.0 & 78.1 & 78.9 & 78.3 \\
GPT-4o
& 0.9 & 25.3 & 15.7 & 82.5 & 87.0 & 64.5 & 71.5 & 69.1 & 68.4 \\
GPT-5 
& 17.2 & 51.2 & 25.6 & 83.4 & 95.6 & 85.2 & 94.9 & 92.0 & 90.7 \\

\midrule
\multicolumn{10}{l}{\textbf{Ours (Based on Qwen3-32B)}} \\
Patent-MAF 
& \textbf{21.1} & \textbf{55.2} & \textbf{32.3} 
& \textbf{88.3} & \textbf{96.4} 
& 84.6 & 90.2 & 85.2 & 86.7 \\
\quad w/o Manager 
& 20.1 & 54.5 & 31.2 & 87.4 & 96.1 & 83.8 & 86.4 & 82.5 & 84.2 \\
\quad w/o Fine-tuning 
& 13.2 & 47.4 & 21.7 & 82.2 & 91.6 & 82.8 & 82.9 & 79.1 & 81.6 \\
\quad w/o Polisher 
& 19.2 & 54.3 & 30.8 & 87.5 & 96.2 & 83.8 & 84.7 & 79.9 & 80.1 \\

\bottomrule
\end{tabular}
}
\caption{Model performance on patent claims. The highest score in each column is marked in \textbf{bold}.}
\label{tab:claim}

\end{table*}

%% file: tabs/tab_datacreate.tex
\begin{table*}[!t]
\centering
\footnotesize

\begin{tabular}{|p{0.95\linewidth}|}
\toprule
\textbf{Task: Pseudo-Disclosure Generation} \\
You will be given a patent title, claims, and full specification. Your task is to extract and rewrite the invention into a clear, human-readable disclosure using plain, non-legal English. Do \textbf{not} copy legal phrasing or claim-style language, and do \textbf{not} introduce any new technical information beyond the original patent. \\
\\
\textbf{Output Structure:} \\
You \textbf{must} generate the disclosure using the following seven invention-centric sections: \\
1. \textbf{Title of the Invention (Layperson Version):} Summarize the invention in one simple sentence. If the original title is technical or obscure, rewrite it into an easy-to-understand name. \\
2. \textbf{Why This Invention Is Needed (Problems in the Prior Art):} Describe the limitations of existing technologies and explain why these problems matter, using simple and intuitive language. \\
3. \textbf{What the Core Idea Is (Essence of the Technical Solution):} Explain the fundamental insight behind the invention. Do not restate patent claims; focus on the underlying idea. \\
4. \textbf{How It Works (Structure, Process, Steps):} Describe the implementation using system structure, workflows, or step-by-step logic, while remaining technically accurate. \\
5. \textbf{What Makes It Different (Novel Features):} Identify key distinguishing features over the prior art and explain why they are important. \\
6. \textbf{What Benefits It Provides (Improvements / Effects):} Clearly state the advantages of the invention compared to conventional approaches. \\
7. \textbf{Optional Variants (Dependent Claims to Plain Language):} Convert dependent claims into human-readable optional improvements, summarized as bullet points (e.g., alternative components, configurable parameters, or variant implementations). \\
\\
\textbf{Output Format:} \\
The output \textbf{must} be a valid JSON object with the following structure: \\
\texttt{\{} \\
\quad \texttt{"title": "...",} \\
\quad \texttt{"problem": "...",} \\
\quad \texttt{"core\_idea": "...",} \\
\quad \texttt{"how\_it\_works": "...",} \\
\quad \texttt{"novelty": "...",} \\
\quad \texttt{"benefits": "...",} \\
\quad \texttt{"optional\_variants": ["...", "...", "..."]} \\
\texttt{\}} \\
\bottomrule
\end{tabular}

\caption{Prompt used to construct pseudo-disclosures for the Dis2Pat dataset.}
\label{tab:disclosure-prompt}
\end{table*}

%% file: tabs/tab_cross.tex
\begin{table*}[!t]
\centering
\footnotesize
\resizebox{.99\textwidth}{!}{
\begin{tabular}{l|ccc|cc|cccc}
\toprule
\multirow{2.7}{*}{\textbf{Model}} & \multicolumn{3}{c|}{\textbf{Text Overlap}} & \multicolumn{2}{c|}{\textbf{Semantic Similarity}} & \multicolumn{4}{c}{\textbf{Content Quality}} \\
\cmidrule(lr){2-4}\cmidrule(lr){5-6}\cmidrule(lr){7-10}
 & \textbf{BLEU} & \textbf{R-1} & \textbf{R-L} & \textbf{BERTScore} & \textbf{BERT-for-Patent} & \textbf{Completeness} & \textbf{Clarity} &  \textbf{Style} & \textbf{Overall} \\
\midrule
\multicolumn{3}{l}{\textbf{Baselines (Text-Only)}} \\

Llama-3.2-90B-Vision 
& 0.5 & 27.3 & 11.8 & 81.5 & 94.5 & 68.7 & 73.6 & 58.1 & 66.8 \\

\midrule
\multicolumn{3}{l}{\textbf{Baselines (Text+Image)}} \\

Llama-3.2-90B-Vision 
& 0.6 & 27.7 & 12.1 & 81.8 & 94.6 & 68.9 & 73.9 & 58.6 & 67.1 \\
\midrule
\multicolumn{3}{l}{\textbf{Ours (Based on Llama-3.2-90B-Vision)}} \\
Patent-MAF
& 0.8 & 28.4 & 13.2 & 82.4 & 95.2 & 72.2 & 76.2 & 64.1 & 70.8 \\

\bottomrule
\end{tabular}
}
\caption{Performance test on patent specification based on Llama-3.2-90B-Vision. }
\label{tab:cross}
\end{table*}

%% file: tabs/tab_speceval.tex
\begin{table*}[!t]
\centering
\footnotesize

\begin{tabular}{|p{0.95\linewidth}|}
\toprule
\textbf{Instructions:} \\
You will receive two patent specifications:
(1) \texttt{gold\_spec}: the gold-standard reference specification, and
(2) \texttt{candidate\_spec}: the specification to be evaluated.
Your task is to evaluate how well the candidate specification matches the gold specification in terms of content coverage, logical clarity, and legal drafting quality. \\
\\
You must assign a score from 0 to 100 for each evaluation dimension, where:
0 indicates extremely poor quality or entirely missing content, and
100 indicates excellent quality fully aligned with professional patent drafting standards. \\
\\
\textbf{Evaluation Criteria:} \\
\textbf{1. Completeness (0--100)} \\
Assess whether the candidate specification adequately covers the technical field, background, problem statement, core technical solution, implementation details, and embodiments present in the gold specification. Scores should reflect missing, incorrect, or sufficiently covered content. \\
\\
\textbf{2. Logical Clarity (0--100)} \\
Evaluate structural coherence, technical flow, causal consistency, definition consistency, and the absence of contradictions, vague references, or logical gaps. \\
\\
\textbf{3. Legal Drafting Quality (0--100)} \\
Assess adherence to patent drafting norms, including avoidance of overly limiting language, consistent terminology usage, and provision of sufficient enablement-level detail. \\
\\
\textbf{Output Format:} \\
Your output \textbf{must} be a valid JSON dictionary with the following exact structure: \\
\texttt{\{} \\
\quad \texttt{"completeness": <score 0--100>,} \\
\quad \texttt{"clarity": <score 0--100>,} \\
\quad \texttt{"legal\_quality": <score 0--100>} \\
\texttt{\}} \\
\bottomrule
\end{tabular}
\caption{LLM-based evaluation prompt used for patent specification quality assessment.}
\label{tab:spec-eval-prompt}
\end{table*}

%% file: tabs/tab_claimeval.tex
\begin{table*}[!t]
\centering
\footnotesize

\begin{tabular}{|p{0.95\linewidth}|}
\toprule
\textbf{Instructions:} \\
You will receive two patent claim sets:
(1) \texttt{Gold Claims}: the gold-standard reference claims, and
(2) \texttt{Candidate Claims}: the claims to be evaluated. \\
\\
\textbf{Role:} \\
You are a Senior Patent Examiner with experience in patent examination. \\
\\
\textbf{Objective:} \\
Evaluate whether the Candidate Claims recite the \emph{same invention} as the Gold Claims, with comparable technical scope and legal rigor.  \\
\\
\textbf{Evaluation Criteria (Score each from 0--100):} \\
\\
\textbf{1. Technical Scope} \\
Assess whether the Candidate Claims capture the core inventive concept of the Gold \emph{independent claims}. \\

\\
\textbf{2. Claim Legality } \\
Evaluate compliance with professional USPTO claim-drafting standards, including: \\
\quad -- Proper independent claim structure (preamble + ``comprising'' + body) \\
\quad -- Correct antecedent basis \\
\quad -- Valid claim dependencies \\
\quad -- Absence of prose, bullet points, or fatal indefiniteness \\

\\
\textbf{3. Logical Consistency \& Clarity} \\
Assess internal coherence, technical plausibility, consistent terminology usage, and absence of contradictions or impossible relationships. \\

\\
\textbf{Output Format:} \\
Your output \textbf{must} be a valid JSON dictionary with the following exact structure: \\
\texttt{\{} \\
\quad \texttt{"scope": <score 0--100>,} \\
\quad \texttt{"legality": <score 0--100>,} \\
\quad \texttt{"clarity": <score 0--100>} \\
\texttt{\}} \\
\bottomrule
\end{tabular}
\caption{LLM-based evaluation prompt used for patent claim quality assessment.}
\label{tab:claim-eval-prompt}
\end{table*}

%% file: tabs/tab_humaneval.tex
\begin{table*}[!t]
\centering
\footnotesize

\begin{tabular}{|p{0.95\linewidth}|}
\toprule
\textbf{Human Evaluation Task: Overall Patent Quality Ranking} \\
\\
\textbf{Evaluation Objective:} \\
You are asked to manually evaluate and compare multiple patent texts generated by different models. Each patent consists of two parts: a \textbf{Specification} and a set of \textbf{Claims}. An officially published or granted patent (\textbf{Gold Patent}) is provided as the upper-bound quality reference. Your goal is to compare two candidate patents against the Gold Patent and rank them by quality. \\
\\
\textbf{Input Materials:} \\
Each evaluation sample includes: \\
-- \textbf{Gold Patent:} The official published or granted patent text, including both claims and specification. \\
-- \textbf{Candidate Patents:} Several complete patent texts generated by different systems. \\
\\
\textbf{Ranking Criteria:} \\
You should rank candidate patents according to the following four criteria, considering the specification and claims jointly: \\
\\
\textbf{1. Completeness} \\
Does the patent sufficiently and fully describe the invention? \\
-- Whether the specification covers the background, problem, technical solution, and embodiments. \\
-- Whether the claims fully reflect the core technical solution described in the specification. \\
-- Whether any critical technical elements necessary for understanding or implementing the invention are missing. \\
\\
\textbf{2. Clarity} \\
Is the patent clear, understandable, and well structured? \\
-- Whether the technical solution is logically organized and easy to follow. \\
-- Whether the claims are clearly stated without ambiguity or redundancy. \\
-- Whether the specification is well organized with coherent paragraphs and sections. \\
\\
\textbf{3. Legality} \\
Does the patent comply with legal and stylistic conventions of patent drafting? \\
-- Whether the claims follow proper legal structure and drafting conventions. \\
-- Whether there is inappropriate de-legalized or conversational language. \\
-- Whether the claims exhibit non-standard or problematic drafting practices (e.g., improper functional language or structural inconsistency). \\
\\
\textbf{4. Consistency} \\
Are the specification and claims mutually consistent and well aligned? \\
-- Whether all technical features recited in the claims are supported by the specification. \\
-- Whether the specification introduces important technical content not reflected in the claims. \\
-- Whether there are contradictions, inconsistent terminology, or logical conflicts between the two parts. \\
\\
\textbf{Output Format (Ranking Only):} \\
Given one Gold Patent and multiple Candidate Patents A, B, output a ranking for each criterion, for example: \\
1. Completeness: A $>$ B \\
2. Clarity: A $<$ B \\
3. Legality: A $=$ B \\
4. Consistency: A $>$ B \\
\bottomrule
\end{tabular}
\caption{Expert human evaluation instructions for overall patent quality ranking.}
\label{tab:human-eval-instructions}
\end{table*}